\documentclass{article}
\usepackage{ICASSP2021,amsmath,graphicx}

\usepackage{booktabs}
\usepackage{multirow} % for table
\usepackage{amssymb} % for \mathbb
\usepackage{xcolor}
\usepackage{hyperref} % for website
\usepackage{float}

\def\x{{\mathbf x}}

\title{Detecting Adversarial Attacks on Audiovisual Speech Recognition}
\name{
    Pingchuan Ma, 
    Stavros Petridis,
    Maja Pantic
}
\address{
Department of Computing, Imperial College London, UK 
}%
\begin{document}
\ninept
\maketitle
\begin{abstract}
Adversarial attacks pose a threat to deep learning models. However, research on adversarial detection methods, especially in the multi-modal domain, is very limited. In this work, we propose an efficient and straightforward detection method based on the temporal correlation between audio and video streams. The main idea is that the correlation between audio and video in adversarial examples will be lower than benign examples due to added adversarial noise. We use the synchronisation confidence score as a proxy for audiovisual correlation and based on it we can detect adversarial attacks. To the best of our knowledge, this is the first work on detection of adversarial attacks on audiovisual speech recognition models. We apply recent adversarial attacks on two audiovisual speech recognition models trained on the GRID and LRW datasets. The experimental results demonstrate that the proposed approach is an effective way for detecting such attacks.
\end{abstract}
\section{Introduction}

\label{sec:intro}
Deep networks achieve state-of-the-art performance on several  tasks such as image classification, image segmentation and face recognition. However, recent studies \cite{szegedy2013intriguing, goodfellow2015explaining} show that such networks are susceptible to adversarial attacks. Given any input $\x$ and a classifier $f(\cdot)$, an adversary tries to carefully construct a sample $\x^{{\rm adv}}$ that is similar to $\x$ but $f(\x) \neq f(\x^{{\rm adv}})$. The adversarial examples are indistinguishable from the original ones but can easily degrade the performance of deep classifiers.

Existing studies on adversarial attacks have mainly focused in the image domain \cite{goodfellow2015explaining, su2019one, athalye2017synthesizing, brendel2018decision}. Recently, adversarial attacks in the audio domain have also been presented \cite{alzantot2017did, carlini2018audio}. One of the most prominent studies is the iterative optimisation-based attack \cite{carlini2018audio}, which directly operates on an audio clip and enables it to be transcribed to any phrase when a perturbation is added. Works on defense approaches against adversarial attacks can be divided into three categories: adversarial training \cite{goodfellow2015explaining}, gradient masking \cite{papernot2016distillation} and input transformation \cite{xu2017feature}. The first one adds adversarial examples in the training set whereas the second one builds a model which does not have useful gradients. Both of them require the model to be retrained, which can be computationally expensive. In contrast, the latter one attempts to defend adversarial attacks by transforming the input.

On the other hand, work on how to detect adversarial attacks is very limited. To the best of our knowledge, the only work in the audio domain was proposed by Yang et al. \cite{yang2019characterizing} and exploits the inherent temporal dependency in audio samples to detect adversarial examples. The main idea is that the transcribed results from an audio sequence and segments extracted from it are consistent in benign examples but not in adversarial ones. In other words, the temporal dependency is not preserved in adversarial sequences.

%\looseness-1
Inspired by the idea of using temporal dependency to detect audio adversarial examples, we propose a simple and efficient detection method against audiovisual adversarial attacks. To the best of our knowledge, this is the first work which presents a detection method of adversarial attacks on audiovisual speech recognition. The key idea is that the audio stream is highly correlated with the video of the face (and especially the mouth region). In case of an adversarial example, the added noise on the audio and video streams is expected to weaken the audiovisual correlation. Hence, we propose the use of audiovisual synchronisation as a proxy to  correlation. In other words, we expect higher synchronisation scores for benign examples and lower scores for adversarial examples. \footnote{ Generated adversarial samples can be seen at \url{https://mpc001.github.io/av_adversarial_examples.html}\label{foot}}

The proposed detection method is tested on speech recognition attacks on models trained on the Lip Reading in the Wild  (LRW) \cite{chung2016lip} and GRID datasets \cite{cooke2006audio}. Our results show that we can detect audiovisual adversarial attacks with high accuracy.

\section{Databases}
\begin{figure*}[htb]
  \centering
  \includegraphics[width=0.9\linewidth]{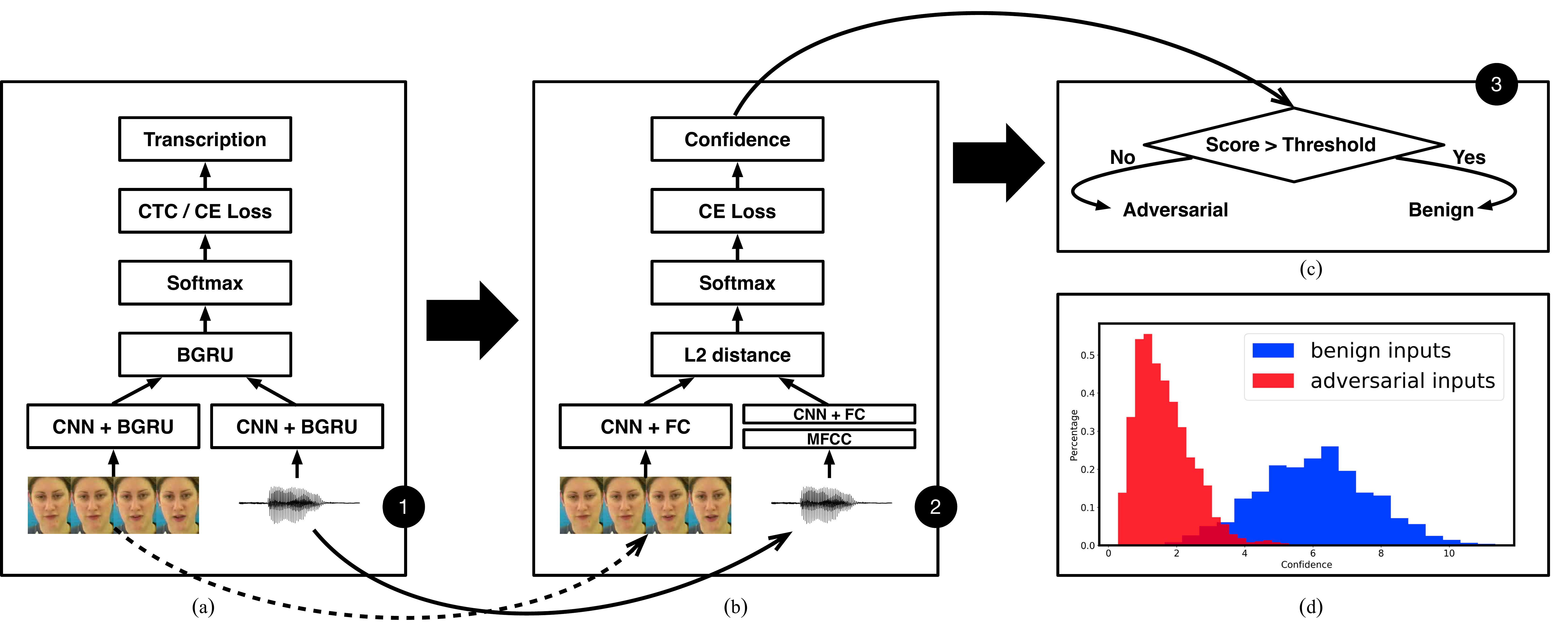}
  \caption{An overview of our proposed detection method. (a) A video and an audio clip are fed to the end-to-end audiovisual speech recognition model. They are also fed to the synchronisation network (b) which estimates a synchronisation confidence score which is used for determining if the audiovisual model has been attacked or not (c). The confidence distribution of 300 adversarial and benign examples from the GRID dataset is shown in (d).}
  \label{fig:overview}
\end{figure*}

For the purposes of this study, we use two audiovisual datasets, the LRW \cite{chung2016lip} and GRID \cite{cooke2006audio} datasets. The LRW dataset  is a large-scale audiovisual dataset consisting of clips from BBC programs. The dataset has 500 isolated words from more than 1000 speakers and contains 488766, 25000, and 25000 examples in the training, validation and test sets, respectively. Each utterance is a short segment with a length of 29 frames (1.16 seconds), where target words are centred in the segment of utterances.

The GRID dataset  consists of 33 speakers and 33000 utterances (1000 per speaker). Each utterance is composed of six words taken from the combination of the following components: \textless command: 4\textgreater \textless colour: 4\textgreater \textless preposition: 4\textgreater \textless letter: 25\textgreater \textless digit: 10\textgreater \textless adverb: 4\textgreater, where the number of choices for each component is indicated in the angle brackets. In this work, we follow the evaluation protocol from \cite{vougioukas2018end} where 16, 7 and 10 subjects are used for training, validation and testing, respectively.

\section{Background}
\subsection{Attacks}
In this study, we consider two attack methods, Fast Gradient Sign Method (FGSM) \cite{goodfellow2015explaining} and the iterative optimisation-based attack \cite{carlini2018audio}. FGSM, which is suitable for attacks on classification models,  computes the gradient with respect to the benign input and each pixel can be updated to maximise the loss. Basic Iterative Method (BIM) \cite{kurakin2016adversarial} is an extended version of FGSM by applying it multiple times with a small step size. Specifically, given a loss function $J(\cdot, \cdot)$ for training the classification model $f(\cdot)$, the adversarial noise $\x^{{\rm adv}}$ is generated as follows: 
\begin{align}
\small
\x_0^{{\rm adv}} &= \x \nonumber \\ \x_{N+1}^{{\rm adv}}&={\rm Clip}_{\x,\epsilon}\{\x_N^{{\rm adv}}+\alpha {\rm sign}(\nabla_\x J(f(\x_N^{{\rm adv}}), y^{{\rm true}})\}
\label{eq:bim}
\end{align}
 
\noindent 
where $\alpha$ is the step size, $\x_N^{{\rm adv}}$ is the adversarial example after $N$-steps of the iterative attack and  $y^{{\rm true}}$ is the true label. After each step, pixel values in the adversarial images $\x^{{\rm adv}}$ are clamped to the range $[\x-\epsilon, \x+\epsilon]$, where $\epsilon$ is the maximum change in each pixel value. This method was proposed for adversarial attacks on images but can also be applied to audio clips by crafting perturbation to the audio input.

The second type of attack \cite{carlini2018audio} has been recently proposed and is suitable for attacks on continuous speech recognition models. Audio adversarial examples can be generated, which can be transcribed to any phrase but sound similar to the benign one. Specifically, the goal of this targeted attack is to seek an adversary input $\x^{{\rm adv}}$, which is very close to the benign input $\x$, but the model decodes it to the target phrase 
$z^{{\rm target}}$. The objective of the attack is the following:
\begin{align}
  \text{minimize} \quad & J(f(\x+\delta), z^{{\rm target}}) \nonumber \\
  \text{such that} \quad & \left\| \delta \right\| < \epsilon
\label{eq:optimisation-based attack}
\end{align}

\noindent
where $\epsilon$ is introduced to limit the maximum change for each audio sample or pixel and $\delta$ is the amount of adversarial noise.

\subsection{Audiovisual Speech Recognition Threat Model}
\label{ssec:AVspeechRecThreatModel}

The architecture is shown in Fig. 1a. We use the end-to-end audiovisual model that was proposed in \cite{ma2019investigating}. The video stream consists of spatiotemporal convolution \cite{stafylakis2017combining}, a modified ResNet18 network and a 2-layer BGRU network whereas the audio stream consists of a 5-layer CNN and a 2-layer BGRU network. These two streams are used for feature extraction from raw modalities. The top two-layer BGRU network further models the temporal dynamics of the concatenated feature.

According to the problem type, two different loss functions are applied for training. The multi-class cross entropy loss, where each input sequence is assigned a single class, is suitable for word-level speech recognition. The CTC loss is used for sentence-level classification. This loss transcribes directly from sequence to sequence when the alignment between inputs and target outputs is unknown. Given an input sequence $\x=(x_1, ..., x_T)$, CTC sums over the probability of all possible alignments to obtain the posterior of the target sequence.

\section{Synchronisation-based Detection Method}
Chung et al. \cite{chung2016out, chung2019perfect} introduced the SyncNet model, which is able to predict the synchronisation error when raw audio and video streams are given. This error is quantified by the synchronisation offset and confidence score. A sliding window approach is used to determine the audiovisual offset. For each 5-frame video window, the offset is found when the distance between the visual features and all audio features in a $\pm$ 1 second range is minimised. The confidence score for a particular offset is defined as the difference between the minimum and the median of the Euclidean distances (computed over all windows). Audio and video are considered perfectly matched if the offset approaches to zero with a high level of confidence score. 

In this work, we aim to explore if such synchronisation is affected by adversarial noise. The detection method is shown in Fig. 1b and 1c. In the detection model, we measure the temporal consistency between the audio and video streams via a model trained for audiovisual synchronisation. For benign audio and video streams, the confidence score should be relatively high since audio and video are aligned and therefore highly synchronised. However, for adversarial audio and video examples, the confidence score is expected to be lower. The added perturbation, which aims to alter the model toward the target transcription, reduces the correlation between the two streams, hence they are less synchronous. Fig. 1d. shows the confidence distribution of 300 benign and adversarial examples from the GRID dataset.

% -- insert figure
\begin{figure*}[tp]
  \centering
  \includegraphics[width=\linewidth]{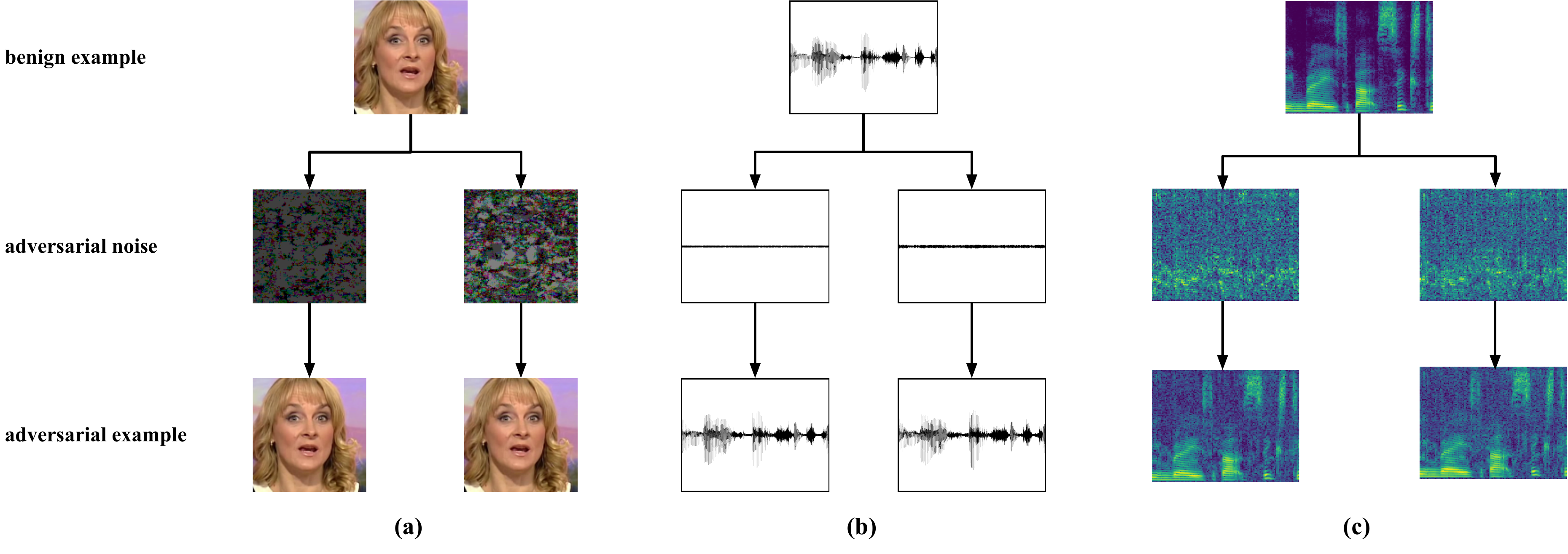}
  \vspace{-.3cm}
  \caption{One example using basic iterative attack on the LRW dataset. Benign examples, adversarial noise examples, and adversarial examples are illustrated from top to bottom. (a) Raw images ($\epsilon^V$=4, $\epsilon^V$=8), (b) audio waveforms ($\epsilon^A$=256, $\epsilon^A$=512), and (c) audio log-spectrum ($\epsilon^A$=256, $\epsilon^A$=512) are presented from left to right. It is noted that the adversarial visual noise has been scaled with a ratio of 64 for a better illustration since the maximum distortion ($\epsilon^V$=8) is 2 pixels.}
  \label{fig:example}
  \vspace{-0.4cm}
\end{figure*}

\section{Experimental Setup}
\subsection{Attacks}

We evaluate our proposed method using two adversarial attacks on both modalities. We assume a white-box scenario, where the parameters of models are known to the attacker.

\textit{Attacks against Word-level Classification}: Attacks such as FGSM and BIM are suitable for word recognition models trained on the LRW dataset. For FGSM,  we consider three values for $\epsilon^A$ used in the audio stream  (256, 512, 1024)  and three values for $\epsilon^V$ for the video stream (4, 8, 16)\footnote{Pixel values are in the range of [0, 255]. Audio samples are in the range of [-32768, 32767].}. For BIM, the step size $\alpha^V$ was set to 1 in the image domain, which means the value of each pixel is changed by 1 at each iteration. The step size $\alpha^A$ in the audio domain is set to 64. We follow the number of iterations setting suggested by \cite{kurakin2016adversarial}, which is selected to be $\min(\epsilon^V+4, 1.25\epsilon^V)$.

\textit{Attacks against Continuous Speech Recognition}: For attacking a speech recognition model trained on GRID we use a recently proposed targeted attack \cite{carlini2018audio}. The maximum distortion allowed as defined by $\epsilon$ (see Eq. \ref{eq:optimisation-based attack}) is limited in $\{256, 512, 1024\}$, $\{4, 8, 16\}$ for audio and video, respectively, and is reduced during iterative optimisation. We implement the attack with 800 iterations. In our studies, 10 random utterances are selected as target utterances. 300 adversarial examples are randomly selected for each target utterance.

\subsection{Evaluation Metrics\label{metrics}}
We use the  Euclidean distance ($L_2$) for measuring the similarity between two images. We also use the $L_\infty$ norm to measure the maximum change per pixel. For audio samples we follow \cite{carlini2018audio} and convert the $L_\infty$ norm to the scale of Decibels ({\rm dB}): ${\rm dB}(\x) = 
\underset{i}{\max}\;20\cdot\log_{10}(x_i)$, where $x_i$ is an arbitrary audio sample point from the audio clip $\x$. The audio distortion is specified as the relative loudness to the benign audio, which can be defined as ${\rm dB}_\x(\delta) = {\rm dB}(\delta) - {\rm dB}(\x)$.

The Area Under the Curve (AUC) score is used for evaluating the detection approach.
We compute the synchronisation confidence score in benign and adversarial examples and by varying the threshold we compute the Receiver Operating Characteristic (ROC) curve.

Finally, in order to compare how this approach would work in a real scenario, we select the threshold (from Fig. 1c) which maximises the average $F_1$ score of adversarial and benign classes on the validation set. Then we use this threshold to compute the average $F_1$ score on the test set.

\section{Results}
\begin{table}
\caption{Results for the proposed adversarial attack detection approach on word recognition models trained on the LRW dataset.$L^V_\infty$ is 1, 2 and 4 pixels when $\epsilon^V$ is 4, 8 and 16, respectively.}\label{result:wordLevel}
\renewcommand{\arraystretch}{1.1}
\begin{center}{\scalebox{0.81}{
\begin{tabular}{|l|c|c|c|c|c|}
\hline
\textbf{Attacks} &\textbf{Top-1} &\multicolumn{2}{c|}{\textbf{Distortion}}& \multicolumn{2}{c|}{\textbf{Measures}}\\\cline{3-6}
\textbf{(Configuration)} &\textbf{Acc.} &$L^V_2$ &$L^A_\infty$\textbf{({\rm dB})} &\textbf{AUC} &$F_1$\\
\hline\hline
FGSM ($\epsilon^A$=1024, $\epsilon^V$=16) &10.40\% &3.46 &-19.26 &0.99 &0.95 \\
FGSM ($\epsilon^A$=512, $\epsilon^V$=16) &21.87\% &3.46 &-25.28 &0.96 &0.89 \\
FGSM ($\epsilon^A$=256, $\epsilon^V$=16) &32.80\% &3.46 &-31.30 &0.90 &0.82 \\ \hline
FGSM ($\epsilon^A$=1024, $\epsilon^V$=8) &12.40\% &1.73 &-19.26 &0.98 &0.94 \\
FGSM ($\epsilon^A$=512, $\epsilon^V$=8) &24.40\% &1.73 &-25.28 &0.94 &0.86 \\
FGSM ($\epsilon^A$=256, $\epsilon^V$=8) &34.73\% &1.73 &-31.30 &0.86 &0.78 \\\hline
FGSM ($\epsilon^A$=1024, $\epsilon^V$=4) &15.20\% &0.87 &-19.26 &0.98 &0.93 \\
FGSM ($\epsilon^A$=512, $\epsilon^V$=4) &27.53\% &0.87 &-25.28 &0.93 &0.85 \\
FGSM ($\epsilon^A$=256, $\epsilon^V$=4) &38.27\% &0.87 &-31.30 &0.83 &0.76 \\ \hline\hline

BIM ($\epsilon^A$=1024, $\epsilon^V$=16) &0.00\% &1.66 &-19.26 &0.90 &0.82 \\
BIM ($\epsilon^A$=512, $\epsilon^V$=16) &0.00\% &1.66 &-25.28 &0.89 &0.81 \\
BIM ($\epsilon^A$=256, $\epsilon^V$=16) &0.00\% &1.70 &-31.30 &0.84 &0.76 \\ \hline
BIM ($\epsilon^A$=1024, $\epsilon^V$=8) &0.00\% &1.07 &-23.34 &0.85 &0.77 \\
BIM ($\epsilon^A$=512, $\epsilon^V$=8) &0.00\% &1.07 &-25.28 &0.85 &0.77 \\
BIM ($\epsilon^A$=256, $\epsilon^V$=8) &0.00\% &1.08 &-31.30 &0.81 &0.74 \\ \hline
BIM ($\epsilon^A$=1024, $\epsilon^V$=4) &0.07\% &0.67 &-29.36 &0.78 &0.72 \\
BIM ($\epsilon^A$=512, $\epsilon^V$=4) &0.07\% &0.67 &-29.36 &0.78 &0.72 \\
BIM ($\epsilon^A$=256, $\epsilon^V$=4) &0.07\% &0.67 &-31.30 &0.77 &0.71 \\\hline
\end{tabular}}}
\end{center}
\vspace{-0.6cm}
\end{table}

\subsection{Word-level Speech Recognition}

\looseness-1
Detection results for attacks on word-level speech recognition are shown in Table \ref{result:wordLevel}. In the presence of adversarial noise, the Top-1 Accuracy drops from 97.20\% \footnote{This is the performance of the model trained on the LRW dataset when benign examples are fed to it.} to below 40\% using FGSM. As $\epsilon^A$ and $\epsilon^V$ increase the accuracy drops (from 38.27\% for the lowest levels of noise to 10.40\% for the highest noise levels). On the other hand, the AUC and F1 scores increase, since the highest levels of noise make detection easier. Similar conclusions can be drawn when BIM is used. Accuracy varies between 0\% and 7\% depending on the noise level, the AUC varies between 0.77 and 0.90 and the F1 scores between 0.71 and 0.82. We should also mention that although  adversarial noise is imperceptible for all values of $\epsilon^V$ it becomes more and more perceptible as $\epsilon^A$ increases.

It is clear from Table \ref{result:wordLevel} that for both types of attacks the distortion is smaller when $\epsilon^A$ and $\epsilon^V$ decrease  and as a consequence detection becomes harder, both AUC and $F_1$ scores go down. However, such attacks are less successful since the classification rate goes up.

We also notice that when the attack is stronger, e.g., BIM is used instead of FSGM,  the classification rate goes down, i.e., the attack is more successful,  and at the same time the distortion ($L^V_2$) becomes smaller.  Consequently, detection becomes more difficult and this is reflected to the lower AUC and $F_1$ scores for BIM than FGSM. 

\begin{table}[t]
\caption{Average results over 10 utterances of the proposed audiovisual synchronisation detection on partially targeted adversarial attacks on continuous speech recognition models trained on GRID. The success rate is the proportion of adversarial examples with WER
less than 50\%. ($\epsilon^A\in\{256, 512, 1024\}$, $\epsilon^V\in\{4, 8\}$)}\label{result:sentenceLevel_differerent configurations}
\renewcommand{\arraystretch}{1.1}
\begin{center}{\scalebox{0.85}{
\begin{tabular}{|c|c|c|ccc|cc|}
\hline
\multicolumn{2}{|c|}{\textbf{Threshold}} &\textbf{Success}&\multicolumn{3}{c|}{\textbf{Distortion}}& \multicolumn{2}{c|}{\textbf{Measures}}\\\cline{4-8} \cline{1-2}
$\epsilon^A$ &$\epsilon^V$ &\textbf{Rate}&$L^V_2$ &$L^V_\infty$ &$L^A_\infty$\textbf{({\rm dB})} &\textbf{AUC} &$F_1$\\
\hline\hline
1024 &8 &100\%& 3.14& 0.019& -43.34& 0.84& 0.75 \\
512 &8 &94\%& 3.38& 0.021& -43.93& 0.84& 0.75 \\
256 &8 &67\%& 3.63& 0.022& -46.77& 0.83& 0.75 \\ \hline
1024 &4 &99\%& 1.54& 0.010& -40.14& 0.79& 0.71 \\
512 &4 &78\%& 0.82& 0.010& -41.14& 0.78& 0.71 \\
256 &4 &42\%& 1.98& 0.012& -45.63& 0.74& 0.68 \\
\hline
\end{tabular}}}
\end{center}
\vspace{-0.5cm}
\end{table}

\subsection{Sentence-level Speech Recognition}
In this section we consider two types of attacks on continuous speech recognition: 1) partially targeted attacks, where the WER between the transcribed result and target phrase is up to 50\%, and 2)  fully targets attacks where the goal of the attack is that the transcribed result is the same as the desired target phrase (WER = 0\%). We also limit the values of $\epsilon^V$ to 4 and 8 since $\epsilon^V$ = 16 results in very perceptible adversarial examples especially in the case of fully targeted attacks.

Average detection results over 10 utterances for partially targeted attacks on sentence-level speech recognition are shown in Table \ref{result:sentenceLevel_differerent configurations}. It is clear that the success rate is pretty high, over 90\% in most cases. Only when $\epsilon^A$ is 256 and $\epsilon^V$ is 4  then the attack is much less successful with a success rate of 42\%. At the same time the detection rates are quite high for most combinations of the two thresholds, varying between 0.74 and 0.84 for AUC and 0.68 to 0.75 for F1 score.

Average detection results over 10 utterances for fully targeted attacks on sentence-level speech recognition are shown in Table \ref{result:sentenceLevel_fully_differerent configurations}. In this case the success rates are much lower than the partially targeted attack due to the difficulty of the task. Relatively high success rates are observed when $\epsilon^V$ is either 4 or 8  and $\epsilon^A$ is 1024 which results in more perceptible adversarial examples. In addition the generated audio and video adversarial examples are more distorted than the ones generated by the partially targeted attacks. In turn, this leads to higher AUC scores, between 0.83 and 0.90, and F1 scores, between 0.77 and 0.82.

Results per sentence for the partially targeted attack \footnote{ bbaazp: bin blue at a zero please, bwbonn: bin white by o nine now, lgwysa: lay green with y seven again, lraces: lay red at c eight soon, pbapoa; place blue at p one again, prbaos: place red by a one soon, prbzts: place red by z two soon, sgifoa: set green in f one again, srixfn: set red in x four now, swipfn: set white in p five now.} when $\epsilon^V$ is 4 and $\epsilon^A$ is 512 are shown
in Table \ref{result:sentenceLevel_50percent_a512v4}. Although the success rates vary a lot (from 62\% to 91\%) depending on the sentence the detection measures AUC and F1 are similar for all sentences. We also observe that the maximum distortions applied to the audio and video signals are similar in most cases.

\begin{table}[t]
\caption{Average results over 10 utterances of the proposed audiovisual synchronisation detection on fully targeted adversarial attacks on continuous speech recognition models trained on GRID. The success rate is the proportion of adversarial examples with WER = 0\%. ($\epsilon^A\in\{256, 512, 1024\}$, $\epsilon^V\in\{4, 8\}$)}\label{result:sentenceLevel_fully_differerent configurations}
\renewcommand{\arraystretch}{1.1}
\begin{center}{\scalebox{0.85}{
\begin{tabular}{|c|c|c|ccc|cc|}
\hline
\multicolumn{2}{|c|}{\textbf{Threshold}} &\textbf{Success}&\multicolumn{3}{c|}{\textbf{Distortion}}& \multicolumn{2}{c|}{\textbf{Measures}}\\\cline{4-8} \cline{1-2}
$\epsilon^A$ &$\epsilon^V$ &\textbf{Rate}&$L^V_2$ &$L^V_\infty$ &$L^A_\infty$\textbf{({\rm dB})} &\textbf{AUC} &$F_1$\\
\hline\hline
1024 &8 &77\%& 3.26& 0.020& -35.22& 0.90& 0.82 \\
512 &8 &36\%& 3.88& 0.024& -39.29& 0.89& 0.81 \\
256 &8 &8\%& 4.15& 0.026& -43.37& 0.87& 0.81 \\ \hline
1024 &4 &66\%& 1.73& 0.011& -34.12& 0.87& 0.78 \\
512 &4 &19\%& 2.13& 0.013& -38.08& 0.84& 0.77 \\
256 &4 &2\%& 2.17& 0.013& -43.85& 0.83& 0.77 \\
\hline
\end{tabular}}}
\end{center}
\vspace{-0.5cm}
\end{table}

\begin{table}[t]
\caption{Results of the proposed audiovisual synchronisation detection on partially targeted adversarial attacks on continuous speech recognition models trained on GRID. The WER between transcribed and target phrases is up to 50\%. The success rate is the proportion of adversarial examples with WER less than 50\%. ($\epsilon^A=512$, $\epsilon^V=4$)}\label{result:sentenceLevel_50percent_a512v4}
\renewcommand{\arraystretch}{1.1}
\begin{center}{\scalebox{0.90}{
\begin{tabular}{|c|c|c|c|c|c|c|}
\hline
\textbf{Target} &\textbf{Success} &\multicolumn{3}{c|}{\textbf{Distortion}}& \multicolumn{2}{c|}{\textbf{Measures}}\\\cline{3-7}
\textbf{Phrase  }&\textbf{Rate}&$L^V_2$ &$L^V_\infty$ &$L^A_\infty$\textbf{({\rm dB})} &\textbf{AUC} &$F_1$\\
\hline
bbaazp& 81\%& 1.842& 0.011& -41.56& 0.78& 0.71 \\
bwbonn& 70\%& 1.877& 0.012& -40.71& 0.79& 0.72 \\
lgwysa& 62\%& 1.956& 0.012& -40.48& 0.80& 0.72 \\
lraces& 78\%& 1.795& 0.011& -41.38& 0.77& 0.70 \\
pbapoa& 91\%& 1.821& 0.011& -41.51& 0.78& 0.71 \\
prbaos& 81\%& 1.734& 0.011& -41.55& 0.77& 0.72 \\
prbzts& 87\%& 1.673& 0.010& -41.87& 0.77& 0.70 \\
sgifoa& 72\%& 1.791& 0.011& -40.97& 0.79& 0.71 \\
srixfn& 76\%& 1.824& 0.011& -40.12& 0.80& 0.70 \\
swipfn& 83\%& 1.700& 0.011& -41.22& 0.78& 0.71 \\
\hline
\end{tabular}}}
\end{center}
\vspace{-0.5cm}
\end{table}

% -------
\section{Conclusion}
In this work, we have investigated the use of audiovisual synchronisation as a detection method of adversarial attacks. We hypothesised that the synchronisation confidence score will be lower in adversarial  than benign examples and demonstrated that this can be used for detecting adversarial attacks. In future work, we would like to investigate more sophisticated approaches for measuring the correlation between audio and visual streams.

% -------------------------------------------------------------------------
\clearpage
\bibliographystyle{IEEEbib}
\bibliography{reference}

\begin{thebibliography}{10}

\bibitem{szegedy2013intriguing}
C.~Szegedy, W.~Zaremba, I.~Sutskever, J.~Bruna, D.~Erhan, I.~Goodfellow, and
  R.~Fergus,
\newblock ``Intriguing properties of neural networks,''
\newblock in {\em Proceedings of International Conference on Learning
  Representations (ICLR)}, 2014.

\bibitem{goodfellow2015explaining}
I.~J Goodfellow, J.~Shlens, and C.~Szegedy,
\newblock ``Explaining and harnessing adversarial examples,''
\newblock in {\em Proceedings of International Conference on Learning
  Representations (ICLR)}, 2015.

\bibitem{su2019one}
J.~Su, D.~V. Vargas, and K.~Sakurai,
\newblock ``One pixel attack for fooling deep neural networks,''
\newblock {\em IEEE Transactions on Evolutionary Computation (TEVC)}, vol. 23,
  pp. 828--841, 2019.

\bibitem{athalye2017synthesizing}
A.~Athalye, L.~Engstrom, A.~Ilyas, and K.~Kwok,
\newblock ``Synthesizing robust adversarial examples,''
\newblock in {\em Proceedings of International Conference on Machine Learning
  (ICML)}, 2018, pp. 284--293.

\bibitem{brendel2018decision}
W.~Brendel, J.~Rauber, and M.~Bethge,
\newblock ``Decision-based adversarial attacks: Reliable attacks against
  black-box machine learning models,''
\newblock in {\em Proceedings of International Conference on Learning
  Representations (ICLR)}, 2018.

\bibitem{alzantot2017did}
M.~Alzantot, B.~Balaji, and M.~Srivastava,
\newblock ``Did you hear that?~adversarial examples against automatic speech
  recognition,''
\newblock in {\em Proceedings of Conference and Workshop on Neural Information
  Processing Systems (NIPSW)}, 2017.

\bibitem{carlini2018audio}
N.~{Carlini} and D.~{Wagner},
\newblock ``Audio adversarial examples: targeted attacks on speech-to-text,''
\newblock in {\em Proceedings of IEEE Security and Privacy (SP) Workshops},
  2018, pp. 1--7.

\bibitem{papernot2016distillation}
N.~Papernot, P.~McDaniel, X.~Wu, S.~Jha, and A.~Swami,
\newblock ``Distillation as a defense to adversarial perturbations against deep
  neural networks,''
\newblock in {\em Proceedings of IEEE Symposium on Security and Privacy (SP)},
  2016, pp. 582--597.

\bibitem{xu2017feature}
W.~Xu, D.~Evans, and Y.~Qi,
\newblock ``Feature squeezing: detecting adversarial examples in deep neural
  networks,''
\newblock in {\em Proceedings of Network and Distributed Systems Security
  (NDSS) Symposium}, 2019.

\bibitem{yang2019characterizing}
Z.~Yang, B.~Li, P.~Chen, and D.~Song,
\newblock ``Characterizing audio adversarial examples using temporal
  dependency,''
\newblock in {\em Proceedings of International Conference on Learning
  Representations (ICLR)}, 2019.

\bibitem{chung2016lip}
J.~S. Chung and A.~Zisserman,
\newblock ``Lip reading in the wild,''
\newblock in {\em Proceedings of Asian Conference on Computer Vision (ACCV)},
  2016, pp. 87--103.

\bibitem{cooke2006audio}
M.~Cooke, J.~Barker, S.~Cunningham, and X.~Shao,
\newblock ``An audio-visual corpus for speech perception and automatic speech
  recognition,''
\newblock {\em The Journal of the Acoustical Society of America}, vol. 120, pp.
  2421--2424, 2006.

\bibitem{vougioukas2018end}
K.~Vougioukas, S.~Petridis, and M.~Pantic,
\newblock ``End-to-end speech-driven facial animation with temporal gans,''
\newblock in {\em Proceedings of British Machine Vision Conference (BMVC)},
  2018.

\bibitem{kurakin2016adversarial}
A.~Kurakin, I.~Goodfellow, and S.~Bengio,
\newblock ``Adversarial examples in the physical world,''
\newblock in {\em Proceedings of International Conference on Learning
  Representations (ICLR) Workshops}, 2017.

\bibitem{ma2019investigating}
P.~Ma, S.~Petridis, and M.~Pantic,
\newblock ``Investigating the {L}ombard effect influence on end-to-end
  audio-visual speech recognition,''
\newblock in {\em Proceedings of Interspeech}, 2019, pp. 4090--4094.

\bibitem{stafylakis2017combining}
T.~Stafylakis and G.~Tzimiropoulos,
\newblock ``Combining residual networks with {LSTM}s for lipreading,''
\newblock in {\em Proceedings of Interspeech}, 2017, vol.~9, pp. 3652--3656.

\bibitem{chung2016out}
J.~S. Chung and A.~Zisserman,
\newblock ``Out of time: automated lip sync in the wild,''
\newblock in {\em Proceedings of Asian Conference on Computer Vision (ACCV)
  Workshops}, 2016, vol. 10117, pp. 251--263.

\bibitem{chung2019perfect}
S.~Chung, J.~Son Chung, and H.~Kang,
\newblock ``Perfect match: Improved cross-modal embeddings for audio-visual
  synchronisation,''
\newblock in {\em Proceedings of IEEE International Conference on Acoustics,
  Speech and Signal Processing (ICASSP)}, 2019, pp. 3965--3969.

\end{thebibliography}

\end{document}